\documentclass[sigconf]{acmart}
\AtBeginDocument{%
  }

\usepackage{enumitem}
 \usepackage{balance}
\setlist{nosep}

\copyrightyear{2025}
\acmYear{2025}
\setcopyright{acmlicensed}\acmConference[MM '25]{Proceedings of the 33rd ACM International Conference on Multimedia}{October 27--31, 2025}{Dublin, Ireland}
\acmBooktitle{Proceedings of the 33rd ACM International Conference on Multimedia (MM '25), October 27--31, 2025, Dublin, Ireland}
\acmDOI{10.1145/3746027.3762038}
\acmISBN{979-8-4007-2035-2/2025/10}

\begin{document}

\title{EVENT-Retriever: Event-Aware Multimodal Image Retrieval for Realistic Captions}

\author{Dinh-Khoi Vo}\orcid{0000-0001-8831-8846}
\affiliation{%
  \institution{University of Science, VNU-HCM}
  \city{Ho Chi Minh}
  \country{Vietnam}
}
\email{vdkhoi@selab.hcmus.edu.vn}

\author{Van-Loc Nguyen}\orcid{0000-0001-9351-3750}
\affiliation{%
  \institution{University of Science, VNU-HCM}
  \city{Ho Chi Minh}
  \country{Vietnam}
}
\email{nvloc@selab.hcmus.edu.vn}

\author{Minh-Triet Tran}\orcid{0000-0003-3046-3041}
\affiliation{%
  \institution{University of Science, VNU-HCM}
  \city{Ho Chi Minh}
  \country{Vietnam}
}
\email{tmtriet@hcmus.edu.vn}

\author{Trung-Nghia Le}\orcid{0000-0003-3046-3041}
\authornote{Corresponding author}
\affiliation{%
  \institution{University of Science, VNU-HCM}
  \city{Ho Chi Minh}
  \country{Vietnam}
}
\email{ltnghia@fit.hcmus.edu.vn}

\renewcommand{\shortauthors}{Dinh-Khoi Vo, Van-Loc Nguyen, Minh-Triet Tran, and Trung-Nghia Le}

\begin{abstract}
Event-based image retrieval from free-form captions presents a significant challenge: models must understand not only visual features but also latent event semantics, context, and real-world knowledge. Conventional vision-language retrieval approaches often fall short when captions describe abstract events, implicit causality, temporal context, or contain long, complex narratives. To tackle these issues, we introduce a multi-stage retrieval framework combining dense article retrieval, event-aware language model reranking, and efficient image collection, followed by caption-guided semantic matching and rank-aware selection. We leverage Qwen3 for article search, Qwen3-Reranker for contextual alignment, and Qwen2-VL for precise image scoring. To further enhance performance and robustness, we fuse outputs from multiple configurations using Reciprocal Rank Fusion (RRF). Our system achieves the \textbf{top-1 score} on the private test set of Track 2 in the EVENTA 2025 Grand Challenge, demonstrating the effectiveness of combining language-based reasoning and multimodal retrieval for complex, real-world image understanding. The code is available at \href{https://github.com/vdkhoi20/EVENT-Retriever}{EVENT-Retriever (GitHub)}.
\end{abstract}




\begin{teaserfigure}
    \centering
    \includegraphics[width=0.85\textwidth]{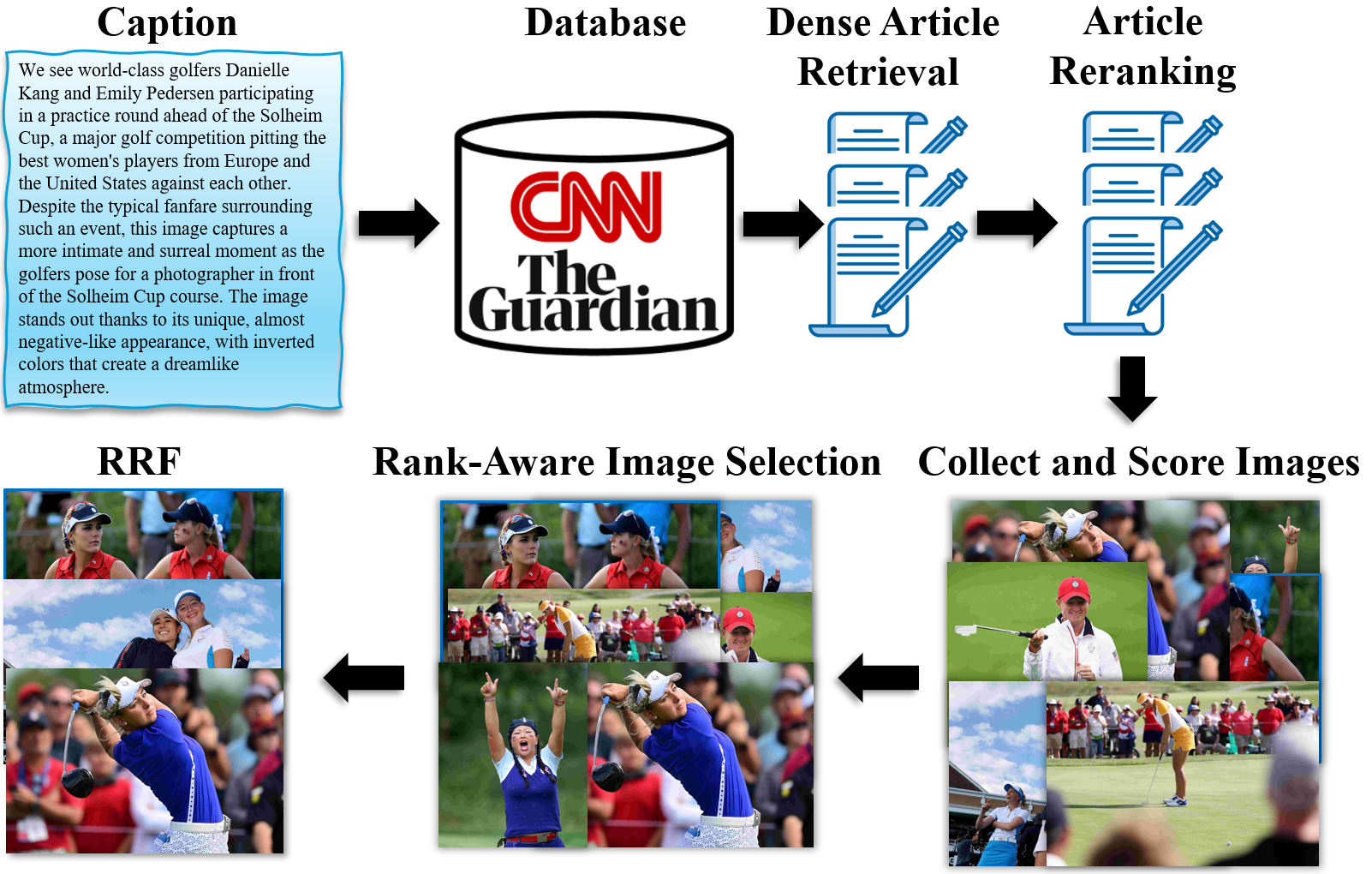}
    \caption{Our event-based retrieval system bridges abstract event descriptions and real-world imagery. Starting from a single caption, we retrieve contextual articles via dense vector search and apply event-aware reranking with a large language model. We then efficiently collect candidate images from the top-ranked articles, score them using cross-modal embeddings, followed by a rank-aware selection strategy that prioritizes images from more relevant articles. Final predictions are refined via Reciprocal Rank Fusion across diverse retrieval configurations. We secured the \textbf{Top-1 score} in Track 2 of the EVENTA 2025 Grand Challenge.}
    \label{fig:pipeline}
\end{teaserfigure}

\maketitle

\section{Introduction}

Image retrieval from natural language input is a fundamental task in computer vision and multimedia research, underpinning a wide range of real-world applications such as content search, e-commerce, and journalism~\cite{liu2021image,dagan2023shop,li2021recent,gudivada1995content}. Traditional retrieval systems excel when queries are short and visually grounded (e.g., “a dog playing with a ball”), but struggle when faced with complex descriptions of real-world events—such as “massive protests following the 2023 election in Argentina” or “wildfires forcing evacuations across southern Europe.” These event-centric queries go beyond surface-level appearances and require models to interpret context, causality, temporal dynamics, and roles of participants~\cite{li2022clip}.

Recent progress in vision-language pretraining, exemplified by CLIP~\cite{radford2021learning}, ALIGN~\cite{jia2021scaling}, BLIP-2~\cite{li2023blip}, and SigLIP~\cite{zhai2023sigmoid}, has significantly advanced text-to-image retrieval through joint embedding spaces trained on large-scale image-caption pairs. These dual-encoder architectures enable zero-shot retrieval by mapping images and texts into a shared representation space, achieving strong results on benchmark datasets such as MSCOCO~\cite{lin2014microsoft} and Flickr30K~\cite{plummer2015flickr30k}. However, they are primarily optimized for short, object-centric captions and often underperform on complex, abstract queries that describe real-world events involving multiple participants, implicit causality, or temporal reasoning. For instance, CLIP is restricted to 77 input tokens, which limits its capacity to represent narrative-rich or multi-sentence captions. Enhancements like SigLIP’s cosine loss or BLIP-2’s integration of frozen vision encoders and LLM decoders improve flexibility but still fall short when handling long-form, compositional semantics and contextual event structures.

Some recent efforts have attempted to address these limitations by incorporating event semantics into the retrieval pipeline. CLIP-Event~\cite{li2022clip} enriches CLIP representations with event structure supervision, using contrastive learning and event graphs to better align textual arguments and visual roles. EventBERT~\cite{zhou2022eventbert} introduces event-type representations for correlation reasoning but focuses primarily on text-only event linking. Liu~et~al.~\cite{liu2022multimedia} propose a unified contrastive framework for multimedia event extraction from news, but their system remains tailored to predefined ontologies and limited-scale datasets. While these approaches highlight the promise of structured event understanding, they are either not designed for open-domain retrieval, lack scalability to long-form documents, or depend on supervised fine-tuning, making them less suitable for real-world, zero-shot image retrieval tasks.

These limitations are magnified in the context of the EVENTA 2025 Grand Challenge (Track 2) ~\cite{eventa25}, which focuses on event-based image retrieval in open-domain, real-world settings. Given a single event-centric caption, systems must retrieve the most relevant image from a large-scale corpus of news articles paired with visual content. The used OpenEvent V1 dataset \cite{nguyen2025openevents} comprises over 202{,}000 articles, each averaging more than 5{,}000 words, and between 1 to over 30 associated images per article. This introduces two key challenges: (1) models must reason over long, context-rich textual inputs that go beyond what is visually observable, and (2) fine-tuning large-scale models on this dataset is computationally prohibitive, necessitating effective zero-shot inference and modular design.

To address these challenges, we propose a scalable, multi-stage retrieval framework powered entirely by Qwen models~\cite{Qwen2.5-VL,Qwen-VL,bai2023qwen,yang2025qwen3}, which demonstrate strong performance across dense retrieval, reranking, and vision-language understanding tasks. The process begins with retrieving semantically relevant articles using dense vector representations generated by {Qwen3-Embedding}~\cite{zhang2025qwen3}, a multilingual embedding model optimized for semantic similarity. Next, we rerank the retrieved articles using {Qwen3-Reranker}~\cite{zhang2025qwen3}, a large-scale causal language model that performs contextual relevance scoring through structured prompting. This step enables document-level reasoning, such as recognizing implicit events or inferring temporal context, to better assess whether an article supports the query.

From the top-ranked articles, we efficiently collect candidate images by traversing the top-ranked articles until at least $I$ valid images are gathered, ensuring they are sourced from a minimum of top-$A$ distinct top-$K$ articles. These collected images are then evaluated using {Qwen2-VL}~\cite{wang2024qwen2}, a vision-language model enhanced with the GME technique~\cite{zhang2024gme}, to compute dense, semantically aligned embeddings for both the caption and each image. This enables robust cross-modal similarity scoring under zero-shot conditions. To finalize the image list, we apply a rank-aware selection strategy that retains a fixed number of top-scoring images per article (e.g., up to three), prioritizing those from higher-ranked articles to ensure both contextual relevance and result diversity.

To further improve robustness and generalization, we employ Reciprocal Rank Fusion (RRF)~\cite{cormack2009reciprocal} to ensemble multiple configurations, including different model scales and limits on the number of images per article. Our system achieves the \textbf{top-1 score} on the private test set of EVENTA Track 2, demonstrating the effectiveness of our modular, reasoning-aware retrieval pipeline under zero-shot settings. The source code is available at \href{https://github.com/Anonymous-Reseacher/EVENT-Retriever}{EVENT-Retriever (GitHub)}.

Our contributions are summarized as follows:
\begin{itemize}
    \item We present a scalable, zero-shot retrieval framework that decomposes event-based image retrieval into four stages: dense article retrieval, contextual reranking, image-level multimodal alignment, and final ensembling via Reciprocal Rank Fusion (RRF).
    
    \item We introduce a prompt-based reranking mechanism using {Qwen3-Reranker}, enabling fine-grained reasoning between narrative-rich captions and long-form article content.

    \item We introduce an efficient image collection strategy that gathers a compact yet diverse pool of candidates from top-ranked articles. These images are then scored using {Qwen2-VL} with {GME} for cross-modal similarity with the query caption, followed by a rank-aware selection strategy that prioritizes high-quality matches from more relevant articles.

    \item We further enhance robustness by ensembling multiple retrieval configurations through {Reciprocal Rank Fusion (RRF)}, improving performance across diverse input cases.

    \item Our approach achieves the \textbf{top-1 score} on the private test set of Track 2 in the \textbf{EVENTA 2025 Grand Challenge}, demonstrating strong generalization in large-scale, real-world event-centric retrieval.
\end{itemize}


\section{Related Work}

\subsection{Vision-Language Retrieval Models}

Early text-to-image retrieval methods used dual-encoder architectures to align image-text pairs in a joint embedding space~\cite{radford2021learning,jia2021scaling}. CLIP and ALIGN achieved strong zero-shot results on datasets like MSCOCO and Flickr30K, but were limited by short, object-centric captions and fixed input sizes (e.g., 77 tokens for CLIP). SigLIP~\cite{zhai2023sigmoid} improved training with cosine contrastive loss but still struggled with abstract or long-form semantics. Fusion-based and generative models such as UNITER~\cite{chen2020uniter}, LXMERT~\cite{tan2019lxmert}, GIT~\cite{wang2022git}, and BLIP-2~\cite{li2023blip} introduced cross-modal attention for richer reasoning, but often hallucinated details and underperformed on open-domain event queries.

\subsection{Multimodal Embedding and Large Language Models}



Recent advances in Multimodal Large Language Models (MLLMs) have extended pretrained language models to vision-language reasoning. Models like GPT-4V~\cite{achiam2023gpt}, LLaVA~\cite{liu2024llavanext,liu2023improvedllava,liu2023llava}, InternVL~\cite{chen2024far}, and MiniCPM-V~\cite{yao2024minicpm} integrate a vision encoder, projection layer, and language model, achieving strong results in instruction following and visual understanding. The Qwen2-VL series~\cite{wang2024qwen2} underpins GME~\cite{zhang2024gme}, a unified multimodal embedder that achieved state-of-the-art performance on the UMRB benchmark~\cite{zhang2025bridging}.

Event-centric retrieval has also been addressed by structure-aware models such as CLIP-Event~\cite{li2022clip}, EventBERT~\cite{zhou2022eventbert}, and contrastive multimedia frameworks~\cite{liu2022multimedia}, though these often face scalability limitations. Meanwhile, text-only retrievers like BGE~\cite{chen2024bge} and E5~\cite{wang2024multilingual} perform well for long-document reranking but require integration with vision-language encoders for multimodal tasks.

\section{Proposed Method}

We propose a scalable and modular framework for event-centric image retrieval from large-scale, weakly aligned multimodal corpora. Our system is designed to meet the key constraints of the EVENTA 2025 challenge, including zero-shot generalization, computational efficiency, and the exclusive use of open-source models. 

The framework consists of four main stages:

\begin{enumerate}
    \item \textbf{Dense Article Retrieval}: We embed all articles offline using Qwen3-Embedding and retrieve top candidates for each query via approximate nearest neighbor search in dense space.
    
    \item \textbf{Article Reranking with Prompted LLM}: Retrieved articles are reranked using Qwen3-Reranker via a structured prompt format to assess whether the article contextually supports the event described in the caption.
    
    \item \textbf{Collect and Score Images via Caption-Guided Matching and Rank-Aware Selection}: We efficiently gather candidate images from relevant articles, compute their similarity to the caption, and apply a rank-aware strategy to select the most appropriate results.

    \item \textbf{Ensembling via Reciprocal Rank Fusion (RRF)}: To improve robustness, we ensemble results from multiple retrieval runs using RRF, giving higher scores to images that consistently appear at top positions across configurations.
\end{enumerate}

All components are built using open-source models from the Qwen family, enabling effective semantic matching and multimodal alignment at scale. The full retrieval pipeline is illustrated in Figure~\ref{fig:pipeline}.

\subsection{Stage 1: Dense Article Retrieval with Qwen3-Embedding}

We begin by retrieving candidate articles that are semantically relevant to the input query. Each article is first formatted by concatenating its \textit{title}, \textit{publication date}, and \textit{content} into a unified document string. These documents are embedded into dense vectors using a multilingual encoder from the Qwen3-Embedding family. All article embeddings are precomputed and stored in a vector database optimized for approximate nearest neighbor (ANN) search using cosine similarity.

At inference time, the event caption is embedded using the same Qwen3-Embedding model. The system retrieves the top-$k$ most similar articles by performing ANN search over the database. This step narrows down the search space to a small candidate pool for subsequent refinement.

\subsection{Stage 2: Article Reranking with Qwen3-Reranker}

To refine the top-$k$ articles retrieved from dense retrieval, we apply a causal language model from the Qwen3-Reranker family. The model is tasked with determining whether a candidate article provides sufficient information to identify the image corresponding to a given event description.

Each input is constructed from three components: a system instruction, a user query–document pair, and an assistant preamble. These are assembled using the model’s native chat format as follows:

\paragraph{System message (prefix):}
This part sets the model's role and constrains the answer format. It is written as:
\begin{quote}
\texttt{<|im\_start|>system}\\
\texttt{Judge whether the Document meets the requirements based on the Query and the Instruct provided. Note that the answer can only be "yes" or "no".}\\
\texttt{<|im\_end|>}
\end{quote}

\paragraph{User message:}
This part encodes the task-specific input and includes three fields: an instruction, a query (caption), and a document (article). It is structured as:
\begin{quote}
\texttt{<|im\_start|>user}\\
\texttt{<Instruct>: Given a caption describing a real-world event, determine if the document provides relevant details to identify the corresponding image. Only answer "yes" or "no".}\\
\texttt{<Query>: A flooded road after heavy rain in Hanoi.}\\
\texttt{<Document>: Title: Torrential Rain Causes Flooding in Hanoi\\
Date: October 14, 2023\\
Content: Several major roads in Hanoi were submerged...}\\
\texttt{<|im\_end|>}
\end{quote}

\paragraph{Assistant preamble (suffix):}
This marks the beginning of the model's output generation:
\begin{quote}
\texttt{<|im\_start|>assistant}\\
\texttt{<think>}\\
\texttt{\ \ \ }\\
\texttt{</think>}\\
\texttt{\ \ \ }
\end{quote}

The entire prompt is tokenized and passed into the language model. At the final token position, we extract the logits over the vocabulary and compute the log-probabilities of the tokens ``yes'' and ``no.'' The probability assigned to ``yes'' is used as a scalar relevance score for the article.

All article–query pairs are processed in batches using multi-GPU inference. The top-$k$ articles with the highest scores are selected for the next stage: image reranking.

\begin{table*}[!t]
\centering
\caption{Comparison of different retrieval pipelines on 1000 queries (caption-to-article). Recall@k scores are reported.}
\label{tab:ablation-study}
\begin{tabular}{ll|rrrrr}
\toprule
\textbf{Embedding} & \textbf{Reranker} &\textbf{Recall@1} & \textbf{Recall@2} & \textbf{Recall@3} & \textbf{Recall@5} & \textbf{Recall@10} \\
\midrule
Bge-m3 && 0.43 & 0.51 & 0.58 & 0.61 & 0.66 \\
Bge-m3 & bge-reranker-v2-m3 & 0.36 & 0.51 & 0.56 & 0.62 & 0.66 \\
Qwen3-Embedding-0.6B & & 0.45 & 0.54 & 0.61 & 0.65 & 0.70 \\
Qwen3-Embedding-0.6B & Qwen3-Reranker-0.6B & 0.51 & 0.63 & 0.65 & 0.68 & 0.70 \\
Qwen3-Embedding-0.6B & Qwen3-Reranker-4B & 0.58 & 0.64 & 0.67 & 0.69 & 0.70 \\
Qwen3-Embedding-0.6B & Qwen3-Reranker-8B & 0.57 & 0.69 & 0.70 & 0.70 & 0.70 \\
Qwen3-Embedding-4B & & 0.52 & 0.61 & 0.65 & 0.68 & 0.71 \\
Qwen3-Embedding-4B & Qwen3-Reranker-0.6B & 0.59 & 0.64 & 0.67 & 0.69 & 0.71 \\
Qwen3-Embedding-4B & Qwen3-Reranker-4B & 0.61 & 0.67 & 0.69 & 0.70 & 0.71 \\
Qwen3-Embedding-4B & Qwen3-Reranker-8B & 0.63 & 0.70 & 0.70 & 0.70 & 0.71 \\
Qwen3-Embedding-8B & & 0.55 & 0.61 & 0.65 & 0.72 & 0.77 \\
Qwen3-Embedding-8B & Qwen3-Reranker-0.6B & 0.60 & 0.65 & 0.68 & 0.70 & 0.77 \\
Qwen3-Embedding-8B & Qwen3-Reranker-4B & 0.64 & 0.69 & 0.73 & 0.74 & 0.77 \\
\textbf{Qwen3-Embedding-8B} & \textbf{Qwen3-Reranker-8B} & \textbf{0.69} & \textbf{0.72} & \textbf{0.75} & \textbf{0.76} & \textbf{0.77} \\
\bottomrule
\end{tabular}
\end{table*}

\subsection{Stage 3: Collect and Score Images via Caption-Guided Matching and Rank-Aware Selection}

After retrieving the top-$K$ articles for each event caption in the previous stage, we efficiently collect a pool of candidate images to evaluate. Specifically, we traverse the top-ranked articles in order, accumulating associated images until we gather at least $I$ valid images from at least $A$ distinct articles among the top-$K$. This early-stopping strategy reduces unnecessary computation while ensuring contextual diversity. Each candidate image is stored along with the rank of its source article. These hyperparameters ($A$, $I$, $K$) are empirically tuned and further discussed in Section~\ref{sec:experiments_caption_article}.

This stage then proceeds in two steps: scoring images based on semantic similarity to the caption, and selecting final results using a rank-aware filtering strategy.

\textit{Step 1: Scoring Images Based on Caption Similarity.}
We compute a dense embedding of the event caption using a vision-language model, and compute embeddings for all collected images in batches. Cosine similarity between caption and image embeddings is then used to score each candidate.

\textit{Step 2: Rank-Aware Image Selection.}
We apply a simple yet effective filtering rule:
\begin{itemize}
    \item From each article (ranked 1 to 10), we keep up to a fixed number of top-scoring images (e.g., 3).
    \item Images are added in article-rank order, giving priority to higher-ranked articles.
    \item If fewer than 10 images are selected, we fill the remaining slots with overflow images or pad with ``\#'' placeholders.
\end{itemize}

This approach ensures that selected images are both semantically aligned with the event and contextually grounded in the most relevant articles.

\subsection{Stage 4: Ensembling via Reciprocal Rank Fusion}

To further improve retrieval robustness, we combine multiple caption-to-image submission outputs using {Reciprocal Rank Fusion (RRF)}~\cite{cormack2009reciprocal}. This approach favors images that consistently appear at high ranks across different configurations.

We collect a set of individual submission files, each containing a list of top-10 image IDs for every query. These submissions are generated from different strategies (e.g., varying the maximum number of images per article or using different model checkpoints). For each query, we aggregate all ranked lists of image IDs and compute an RRF score for each image using the formula $\text{score}(i) = \sum_{r \in \mathcal{R}_i} \frac{1}{k + r}$, where $\mathcal{R}_i$ is the set of ranks of image $i$ across all runs, and $k$ is a smoothing constant (we use $k=60$).

After computing RRF scores, we select the top-$K$ images per query based on their fused rankings. If fewer than 10 unique images are available, we pad the output with a special placeholder (e.g., ``\#'') to maintain a fixed-length output.

This ensembling step increases recall and balances precision across diverse retrieval pipelines. It also reduces the impact of noise or suboptimal choices in any single configuration.

\section{Experiments}

\subsection{Environmental Settings}

All experiments were conducted on the OpenEvents V1 dataset~\cite{nguyen2025openevents}. We evaluated the methods using the official metrics defined by the EVENTA 2025 Challenge\footnote{\url{https://ltnghia.github.io/eventa/eventa-2025/track2}} \cite{eventa25}.

\subsection{Caption-to-Article Retrieval Performance}
\label{sec:experiments_caption_article}

We evaluated our article retrieval stage using $1000$ representative event captions from the training set. As shown in Table \ref{tab:ablation-study}, Qwen3-Embedding~\cite{zhang2025qwen3} models consistently outperformed the BGE-m3~\cite{multi2024m3} baseline.  Specifically, the Qwen3-Embedding-8B combined with Qwen3-Reranker-8B achieved the highest performance, with Recall@1 of 0.69 and Recall@3 of 0.75, compared to Recall@1 of 0.43 for BGE-m3. We observed substantial gains as model capacity increased, and reranking with larger Qwen3-Reranker variants further improved retrieval accuracy.

These results demonstrate the importance of combining strong dense retrieval with powerful language model reranking, particularly for handling long, complex event queries. Based on these findings, we configure the subsequent image collection step to stop early once at least 10 candidate images are collected from at least 3 of the top-10 ranked articles. This strategy provides a diverse and contextually relevant image pool for downstream retrieval.

\subsection{Caption-to-Image Retrieval Performance}
\label{sec:caption-to-image}

We further assess the performance of various vision-language models on a controlled caption-to-image retrieval task. Each query consists of a ground-truth caption and its corresponding article ID, from which we extract all images associated with that article (typically ranging from 1 to 30). The goal is to rank the ground-truth image higher than others within its article context.

We evaluate on the full training set, covering over 20,000 such queries, and report Recall@$k$ scores in Table~\ref{tab:caption-to-image}. CLIP-based models such as \textit{clip-vit-large-patch14}~\cite{radford2021learning} and \textit{siglip-so400m-patch14-384}~\cite{zhai2023sigmoid} achieve strong performance but are inherently limited by the 77-token input constraint, making them less effective for long, narrative-rich captions. BLIP2's ~\cite{li2023blip} performance remains modest due to reliance on generative alignment. In contrast, our approach based on \textit{gme-Qwen2-VL-2B-Instruct}~\cite{zhang2024gme}, which supports fused-modal embeddings with long-text input, delivers the best performance across all Recall@$k$ metrics. This validates the advantage of GME in scenarios requiring fine-grained, long-form semantic alignment between image and caption.

\subsection{Public Test Set Evaluation and Comparative Analysis}

To evaluate the robustness and generalization of our approach, we submitted multiple system variants to the EVENTA 2025 public leaderboard. Table~\ref{tab:public-test-combined} presents the top-ranking systems, including leading external teams and all our internal configurations based on Qwen and BGE.

Our best-performing configuration—\textbf{Qwen3-Embedding-8B + Qwen3-Reranker-8B}—achieved an overall score of {0.5378}, ranking {3rd} on the public leaderboard and coming within just 0.035 points of the top system. This demonstrates the strong generalization ability of Qwen-based models in handling real-world, event-centric retrieval under zero-shot conditions.

\begin{table}[!t]
\centering
\caption{Caption-to-image retrieval results evaluated over the full training set (20K+ queries). For each query, only images within the ground-truth article are considered.}
\label{tab:caption-to-image}
\resizebox{\linewidth}{!}{
\begin{tabular}{lrrrrr}
\toprule
\textbf{Model} & \textbf{R@1} & \textbf{R@3} & \textbf{R@5} & \textbf{R@10} \\
\midrule
{Blip2-itm-vit-g}                 & 0.3703 & 0.6377 & 0.7555 & 0.8932 \\
{clip-vit-large-patch14}         & 0.7844  & 0.9261 & 0.9671 & 0.9920 \\
{Siglip-so400m-patch14-384}      & 0.7565  & 0.9261 & 0.9581 & 0.9890 \\
\textbf{{gme-Qwen2-VL-2B-Instruct}} & \textbf{0.8543} & \textbf{0.9641} & \textbf{0.9880} & \textbf{1.0000} \\
\bottomrule
\end{tabular}
}
\end{table}

\begin{table*}[!t]
\centering
\caption{Leaderboard comparison on the EVENTA 2025 public test set. Our Qwen-based systems significantly outperform BGE and closely match top external teams.}
\label{tab:public-test-combined}
\begin{tabular}{lcccccc}
\toprule
\textbf{Participant} & \textbf{mAP} & \textbf{MRR} & \textbf{R@1} & \textbf{R@5} & \textbf{R@10} & \textbf{Overall Score} \\
\midrule
Sharingan Retrievers & 0.559 & 0.559 & 0.454 & 0.702 & 0.760 & 0.5727 \\
23trinitrotolue & 0.539 & 0.539 & 0.448 & 0.666 & 0.704 & 0.5516 \\
Ours (Qwen3-Embed-8B + Qwen3-Reranker-8B + gme-Qwen2-VL-2B-Instruct) & 0.525 & 0.525 & 0.426 & 0.657 & 0.720 & 0.5378 \\
Ours (Qwen3-Embed-4B + Qwen3-Reranker-4B + gme-Qwen2-VL-2B-Instruct) & 0.507 & 0.507 & 0.410 & 0.639 & 0.696 & 0.5200 \\
Re: Zero Slavery & 0.489 & 0.489 & 0.380 & 0.643 & 0.697 & 0.5005 \\
chisboiz111 & 0.489 & 0.489 & 0.380 & 0.643 & 0.697 & 0.5005 \\
\textbf{Ours (BGE-m3 + BGE-Reranker-v2-m3+ + gme-Qwen2-VL-2B-Instruct)} & 0.420 & 0.420 & 0.331 & 0.533 & 0.610 & 0.4311 \\
\bottomrule
\end{tabular}
\end{table*}

\begin{table*}[!t]
\centering
\caption{Final results on the EVENTA 2025 private leaderboard. The Reciprocal Rank Fusion (RRF) ensemble outperforms all participants, validating its robustness.}
\label{tab:final-results}
\begin{tabular}{lcccccc}
\toprule
\textbf{Participant} & \textbf{mAP} & \textbf{MRR} & \textbf{R@1} & \textbf{R@5} & \textbf{R@10} & \textbf{Overall Score} \\
\midrule
\textbf{Ours (Reciprocal Rank Fusion)} & \textbf{0.563} & \textbf{0.563} & \textbf{0.469} & 0.690 & 0.744 & \textbf{0.5766} \\
23trinitrotolue & 0.558 & 0.558 & 0.456 & \textbf{0.698} & \textbf{0.762} & 0.5722 \\
Ours (Qwen3-Embed-8B + Reranker-8B + gme-Qwen2-VL-2B-Instruct) & 0.552 & 0.552 & 0.445 & 0.675 & 0.733 & 0.5712 \\
Ours (Qwen3-Embed-4B + Reranker-4B + gme-Qwen2-VL-2B-Instruct) & 0.546 & 0.546 & 0.438 & 0.669 & 0.728 & 0.5672 \\
LastSong & 0.549 & 0.549 & 0.449 & 0.695 & 0.738 & 0.5635 \\
\bottomrule
\end{tabular}
\end{table*}

Comparing our internal submissions, we observe that both Qwen configurations substantially outperform the BGE baseline across all metrics. The Qwen3-4B setup already provides a significant gain over BGE (+9.7\% in mAP, +7.9\% in R@1), and the 8B variant further improves these results. These findings suggest that:
\begin{itemize}
    \item Larger Qwen models capture richer semantic and contextual cues crucial for aligning narrative captions with relevant articles and images.
    \item BGE, while lightweight and efficient, lacks the depth required for deep event understanding and complex retrieval.
\end{itemize}

These results justify our design choice of adopting Qwen3-based architectures as the backbone for both dense retrieval and reranking stages in our pipeline.

\subsection{Final Results and Insights}

The final evaluation on the private test set of EVENTA 2025 confirms the effectiveness of our modular retrieval pipeline. Table~\ref{tab:final-results} reports the top submissions, including our best-performing systems with and without ensembling.

Our final ensemble submission, constructed using {Reciprocal Rank Fusion (RRF)} across multiple high-performing runs, achieved an overall score of {0.5766}, outperforming both our strongest individual model (\textit{Qwen3-8B + Reranker-8B}) and all external submissions. Notably, the RRF result delivered the best Recall@1, indicating stronger precision in top-ranked predictions.

RRF combines diverse retrieval strategies by aggregating rankings rather than raw scores, allowing the system to benefit from complementary strengths of each configuration while mitigating their individual weaknesses. For instance, models tuned for aggressive recall or conservative precision can both contribute useful candidates that would otherwise be missed in a single-pass pipeline.

These results validate the importance of ensembling in high-stakes retrieval settings. While large-scale Qwen models already perform competitively, their combination through a lightweight yet effective strategy like RRF provides a consistent and robust boost to final performance. This supports our system design philosophy: modularity, compositionality, and diversity-aware ensembling for real-world event-based image retrieval.

\section{Limitations and Discussion}

While our system achieved top-tier performance on the EVENTA 2025 challenge, several limitations remained. Due to the limited timeframe of the competition, we were unable to exhaustively explore the full space of available vision-language models. Many recent models—such as those based on diffusion, hybrid transformers, or newer generative rerankers—were not tested. Furthermore, we did not leverage the provided training set for supervised fine-tuning or few-shot adaptation, as our approach focused on zero-shot generalization to align with challenge constraints and resource limitations. With more time and compute, integrating instruction tuning or contrastive alignment from curated training examples could further improve both article and image retrieval. Lastly, our current pipeline processes each caption independently and handles article and image retrieval in separate stages. A promising future direction would be to explore end-to-end training or joint modeling of captions directly to (article content + associated images), enabling tighter multimodal alignment and potentially stronger semantic coherence across stages.

\section{Conclusion}

We proposed a scalable and modular retrieval framework for event-based image retrieval over large, weakly aligned multimodal corpora. Our method combined dense article retrieval, prompt-based reranking using large language models, and image reranking via multimodal similarity, all powered by Qwen open-source models. To improve robustness and reduce reliance on any single configuration, we applied Reciprocal Rank Fusion to ensemble multiple retrieval outputs. This approach allowed the system to reason over event structure, temporal context, and visual grounding—all under a zero-shot setting without any fine-tuning. Our solution ranked first in Track 2 of the EVENTA 2025 Grand Challenge, highlighting the strength of unified vision-language architectures and structured retrieval pipelines for complex real-world tasks.


\begin{acks}


 This research is funded by Vietnam National University - Ho Chi Minh City (VNU-HCM) under Grant Number C2024-18-25. 
 
\end{acks}

\bibliographystyle{ACM-Reference-Format}
\balance
\bibliography{sample-base}

\end{document}